\pdfoutput=1

\documentclass[11pt]{article}

\usepackage{acl}

\usepackage{times}
\usepackage{latexsym}

\usepackage[T1]{fontenc}

\usepackage[utf8]{inputenc}

\usepackage{microtype}

\usepackage{inconsolata}
\usepackage{microtype}
\usepackage{graphicx}
\usepackage{import}
\usepackage{adjustbox}
\usepackage{layout}
\usepackage{tabularx, makecell}
\usepackage{booktabs}
\usepackage{mathrsfs}
\usepackage{amssymb} 
\usepackage{url}
\usepackage{hyperref}
\usepackage{graphicx}
\usepackage{xspace,paralist}
\usepackage{times,latexsym}
\usepackage{amsmath}
\usepackage{appendix}
\usepackage{comment} 
\usepackage{enumitem}
\usepackage{makecell}
\usepackage{multirow}
\usepackage{xcolor}
\usepackage{arydshln}
\usepackage{cleveref}
\usepackage{todonotes}
\usepackage{longtable,supertabular}
\usepackage{amssymb}
\usepackage{multirow}
\usepackage{array}
\usepackage{pifont}
\usepackage{CJKutf8}

\newcommand{\tabref}[2][]{Table#1~\ref{#2}\xspace}

\newcommand{\appref}[1]{Appendix~\ref{#1}\xspace}

\NewDocumentCommand{\revanth}
{ mO{} }{\textcolor{blue}{\textsuperscript{\textit{Revanth}}\textsf{\textbf{\small[#1]}}}}
\title{GenAI Content Detection Task 1: English and Multilingual Machine-Generated Text Detection: AI vs. Human}

\author{Yuxia Wang\textsuperscript{1},
Artem Shelmanov\textsuperscript{1},
Jonibek Mansurov\textsuperscript{1},
Akim Tsvigun\textsuperscript{2,3}, 
\\\bf
Vladislav Mikhailov\textsuperscript{4}, 
Rui Xing\textsuperscript{1}, 
Zhuohan Xie\textsuperscript{1}, 
Jiahui Geng\textsuperscript{1}, 
Giovanni Puccetti\textsuperscript{5}, 
\\\bf
Ekaterina Artemova\textsuperscript{6}, 
Jinyan Su\textsuperscript{1, 10}, 
Minh Ngoc Ta\textsuperscript{9}, 
Mervat Abassy\textsuperscript{13},
Kareem Elozeiri\textsuperscript{11}, 
\\\bf
Saad El Dine Ahmed\textsuperscript{13}, 
Maiya Goloburda\textsuperscript{1}, 
Tarek Mahmoud\textsuperscript{1},
Raj Vardhan Tomar\textsuperscript{14},
\\\bf
Alexander Aziz\textsuperscript{15}, 
Nurkhan Laiyk\textsuperscript{1}, 
Osama Mohammed Afzal\textsuperscript{1}, 
Ryuto Koike\textsuperscript{7},
\\\bf
Masahiro Kaneko\textsuperscript{1,7}, 
Alham Fikri Aji\textsuperscript{1}, 
Nizar Habash\textsuperscript{1,8}, 
Iryna Gurevych\textsuperscript{1,12}, 
Preslav Nakov\textsuperscript{1}
\\
\textsuperscript{1}MBZUAI \quad
\textsuperscript{2}Nebius AI \quad 
\textsuperscript{3}KU Leuven \quad
\textsuperscript{4}University of Oslo 
\textsuperscript{5}ISTI-CNR\quad
\textsuperscript{6}Toloka AI
\\
\textsuperscript{7}Institute of Science Tokyo \quad
\textsuperscript{8}New York University Abu Dhabi
\\
\textsuperscript{9}BKAI Research Center, Hanoi University of Science and Technology \quad
\textsuperscript{10}Cornell University
\\
\textsuperscript{11}Zewail City of Science and Technology \quad
\textsuperscript{12}TU Darmstadt \quad
\textsuperscript{13}Alexandria University
\\
\textsuperscript{14}Cluster Innovation Center, University of Delhi \quad
\textsuperscript{15}University of Florida
\\
\href{yuxia.wang@mbzuai.ac.ae}{\{yuxia.wang, artem.shelmanov, preslav.nakov\}@mbzuai.ac.ae}
}

\begin{document}
\maketitle
\begin{abstract}
We present the GenAI Content Detection Task~1 -- a shared task on binary machine generated text detection, conducted as a part of the GenAI workshop at COLING 2025. The task consists of two subtasks: Monolingual (English) and Multilingual. The shared task attracted many participants: 36 teams made official submissions to the Monolingual subtask during the test phase and 26 teams -- to the Multilingual. 
We provide a comprehensive overview of the data, a summary of the results -- including system rankings and performance scores -- detailed descriptions of the participating systems, and an in-depth analysis of submissions.\footnote{\url{https://github.com/mbzuai-nlp/COLING-2025-Workshop-on-MGT-Detection-Task1}}
\end{abstract}

\section{Introduction}
The success and popularity of Large Language Models (LLMs) have led to the proliferation of generative artificial intelligence (GenAI) content, which is now widely applied across numerous aspects of daily life. 
However, this widespread adoption has brought several concerns to light, including challenges to the integrity of student assignments and the potential for fabricated content to mislead individuals~\cite{wang-etal-2024-m4}.
As generative LLMs continue to advance rapidly, it is becoming increasingly difficult for humans to distinguish machine-generated content from authentic human-authored text. Consequently, developing effective methods to address these challenges is crucial. To this end, we propose a GenAI content detection task, with Task 1 focusing specifically on the detection of machine-generated text in both English and multilingual contexts. This task is the continuation of SemEval Shared Task 8~\cite{wang2024semeval}. 
The new task introduces a broader range of languages and domains while incorporating updated generators that leverage the latest LLMs.

The task consists of two subtasks: \textbf{Monolingual (English) subtask A} and \textbf{Multilingual subtask~B}. The data for the shared task covers various domains and LLM generators. The data for English subtask covers diverse domains, including peer reviews, student essays, scientific papers, news articles, social media, emails, speech content and so on, similar for multilingual subtask data, with the test set involving more than 8 domains.
To construct the data for the shared task, we produced machine-generated texts (MGTs), using state-of-the-art LLMs, including GPT-4/4o, Mistral~\cite{jiang2023mistral}, Llama-3.1~\cite{dubey2024llama}, Vikhr-Nemo~\cite{nikolich-etal-2024-vikhr}, Qwen-2~\cite{yang2024qwen2}, etc. Multilingual subtask data encompasses 21 unique languages. 

The task attracted 36 participants who made official submissions during the test phase for the monolingual subtask A and 27 participants who made official submissions to the multilingual subtask B. 
\section{Related Work}
This section discusses prior work about machine-generated text detection methods, datasets and shared tasks.

\subsection{Detection Methods}
There are mainly two commonly used approaches for detecting machine-generated text, training-free and training-based. 
Training-free detection methods leverage statistical characteristics of texts for identifying MGTs \cite{solaiman2019release, gehrmann2019gltr}. Various features have been explored, such as perplexity \cite{vasilatos2023howkgpt}, perplexity curvature \cite{mitchell2023detectgpt}, log rank \cite{su2023detectllm}, intrinsic dimensionality \cite{tulchinskii2024intrinsic} and N-gram analysis \cite{yang2023dna}. Revise-Detect hypothesizes that machine-generated texts would be edited less by LLMs than human-written texts \cite{zhu2023beat}. Binoculars \cite{hans2024spotting} employs two LLMs to calculate the ratio of perplexity to cross-perplexity, assessing how one LLM responds to the next token predictions of another. 
Training based detectors typically fine-tune a pre-trained model for binary classification  \cite{yu2023gpt,zhan2023g3detector}, utilizing techniques such as adversarial training \cite{hu2023radar} and abstention \cite{tian2023multiscale}. \citet{verma2023ghostbuster} fine-tune a linear classifier on top of the learned representations.

\subsection{Datasets}
There are many efforts in detecting machine-generated text benchmarks. HC3 \cite{guo2023close} contains both Chinese and English text from ChatGPT. Other datasets such as MGTBench \cite{he2023mgtbench}, ArguGPT \cite{liu2023argugpt} and DeepfakeTextDetect~\cite{li2023deepfake} consider texts generated by various LLMs. 
M4 and M4GT-Bench \cite{wang-etal-2024-m4, wang2024m4gtbench} are two comprehensive datasets covering multiple domains, languages and generators. 
MULTITuDE \cite{macko-etal-2023-multitude} includes texts in 11 languages, while the MAiDE-up dataset \cite{ignat2024maide} focuses on hotel reviews generated in 10 languages by GPT-4. MultiSocial \cite{macko2024multisocial} benchmarks MGT detection in the social media domain for 22 languages and 5 social media platforms.

\subsection{Shared Tasks}
Several shared tasks have been organized to address the problem of detecting machine-generated texts. 
\textit{2023 ALTA shared task} \cite{molla2023overview} focuses specifically on identifying GPT-generated texts. 
\textit{DAGPap22 shared task } \cite{chamezopoulos2024overview} targets the detection of machine-generated scientific papers. 
\textit{SemEval 2024 shared task 8} \cite{wang2024semeval} introduced four subtasks: monolingual and multilingual binary classification (whether the text is generated by machine or written by human), multi-way classification distinguishing different generators, and human-machine text boundary detection, attracting participation from hundreds of teams.

There has been growing interest in detecting machine-generated text in non-English languages, such as Russian in \textit{RuATD Shared task 2022} \cite{shamardina2022findings,shamardinacoat}, Spanish in \textit{IberLEF 2023}  \cite{sarvazyan2023overview}, and Dutch in \textit{CLIN33} \cite{fivez2024clin33}. The multilingual detection task on SemEval-2024 Task 8 \cite{wang2024semeval} covers 9 languages, utilizing the M4GT-Bench dataset~\cite{wang-etal-2024-m4gt}.
\section{Shared Task Description}

\subsection{Overview}

The shared task was conducted in two phases: the development phase August 27, 2024 -- October 29, 2024 and the test phase October 30 -- November 4, 2024. During the training phase, the participants were given access to the texts and labels of the training and validation subsets, as well as to the texts of the dev-test subset. 
The dev-test set was made available to participants to evaluate the generalization capabilities of their detectors on distinct data during the development phase.


After the start of the test phase, we opened the labels of the dev-test and provided access to the texts of the test subset with a limited number of submission attempts to prevent leakage. After the finish of the test phase, we have released the labels of the test set, so the participants could perform some ablation studies. 

As per the rules of the Task, participants were required to use only the data provided by the organizers to develop their models and were prohibited from utilizing any additional training data.

\subsection{Datasets}
\label{sec:dataset}


The data for the Task is split into four subsets: training, development, dev-test, and test. Texts and labels for all subsets are publicly available at \href{https://github.com/mbzuai-nlp/COLING-2025-Workshop-on-MGT-Detection-Task1}{Github repository}.
Tables~\ref{tab:english-data} and \ref{tab:multilingual-data} present the descriptive statistics of the data. 

\begin{table*}[t!]
    \centering
    \small
    \begin{tabular}{l|c|lcc|rrr|r}
    \toprule
    \textbf{Split} & \textbf{Source} & \textbf{Data License} & \textbf{\#Generators} & \textbf{\#Domains} & \textbf{Human} & \textbf{MGT} & \textbf{H+M} & \textbf{Total}  \\
    \midrule
    \multirow{3}{*}{Train} & HC3 & CC BY-SA-4.0 & 1 & 5 & 39,140 & 18,671 & 57,811 & \multirow{3}{*}{610,767} \\
    & M4GT & CC BY-SA-4.0 & 14 & 6 & 86,782 & 181,081 & 267,863 & \\
    & MAGE & Apache-2.0 & 27 & 14 & 103,000 & 182,093 & 285,093 & \\
    \midrule
    \multirow{3}{*}{Dev} & HC3 & CC BY-SA-4.0 & 1 & 5 & 16,855 & 7,917 & 24,772 & \multirow{3}{*}{261,758} \\
    & M4GT & CC BY-SA-4.0 & 14 & 6 & 37,220 & 77,267 & 114,487 & \\
    & MAGE & Apache-2.0 & 27 & 14 & 44,253 & 78,246 & 122,499 & \\
    \midrule
    \multirow{2}{*}{Dev-test} & RAID & MIT & 0 & -- & 13,371 & 0 & 13,371 & \multirow{2}{*}{32,557}\\
    &  LLM-DetectAIve & CC BY-SA-4.0 & 5 & -- & 0 & 19,186 & 19,186 & \\
    \midrule
    \multirow{5}{*}{Test} & CUDRT & CC BY-SA-4.0 & 6 & 6 & 12,287 & 10,691 & 22,978 & \multirow{5}{*}{73,941} \\
    & IELTS & Apache-2.0 & 2 & 1 & 11,382 & 13,318 & 24,700 & \\
    & NLPeer & Apache-2.0 & 1 & 1 & 5,326 & 5,376 & 10,702\\
    & PeerSum & Apache-2.0 & 2 & 1 & 5,080 & 6,995 & 12,075 & \\
    & MixSet & CC BY-SA-4.0 & 7 & 9 & 600 & 2,886 & 3,486 & \\
    \midrule
    \textbf{Total} &  &  &  & & 375,296 & 603,727 & 979,023 & \textbf{979,023}\\
    \bottomrule
    \end{tabular}
    \caption{\textbf{English subtask} statistical information of training, development, dev-test, and test sets.}
    \label{tab:english-data}
\end{table*}

\begin{table*}[t!]
    \centering
    \small
   \resizebox{\textwidth}{!}{
    \begin{tabular}{l|c|llcc|rrr|r}
    \toprule
    \textbf{Split} & \textbf{Source} & \textbf{Data License} & \textbf{Lang} & \textbf{\#Generators} & \textbf{\#Domains} & \textbf{Human} & \textbf{MGT} & \textbf{H+M} & \textbf{Total}  \\
    \midrule
    \multirow{3}{*}{Train} & HC3 & CC BY-SA-4.0 & zh, en & 1 & 9 & 54,655 & 30,670 & 85,325 & \multirow{3}{*}{674,083}\\
    & M4GT & CC BY-SA-4.0 & 9 & 16 & 13 & 100,359 & 203,525 & 303,884 & \\
    & MAGE & Apache-2.0 & en & 27 & 14 & 102,954 & 181,920 & 284,874 & \\
    \midrule
    \multirow{3}{*}{Dev} & HC3 & CC BY-SA-4.0 & zh, en & 1 & 9 & 22,981 & 12,718 & 35,699 & \multirow{3}{*}{288,894}\\
    & M4GT & CC BY-SA-4.0 & 9 & 16 & 13 & 42,886 & 87,591 & 130,477 & \\
    & MAGE & Apache-2.0 & en & 27 & 14 & 44,299 & 78,419 & 122,718 & \\
    \midrule
    Dev-test & MULTITuDE & GPL-3.0 & 11 & 8 & -- & 7,992 & 66,089 & 74,081 & 74,081 \\
    \midrule
    Test & 29 sources & -- & 15 & 19 & -- & 73,634 & 77,791 & 151,425 & 151,425 \\
    \midrule
    \textbf{Total} &  &  &  & & & 449,760 & 738,723 & 1,188,483 & \textbf{1,188,483}\\
    \bottomrule
    \end{tabular}
   }
    \caption{\textbf{Multilingual subtask} statistics of training, development, dev-test, and test sets. M4GT includes 9 languages: en, de, id, it, zh, bg, ar, ur, ru. MULTITuDE includes 11 languages: de, en, uk, es, nl, ca, ru, pt, ar, zh, cs.}
    \label{tab:multilingual-data}
\end{table*}

\subsubsection{Training and Development Sets}
The training data for both English and multilingual subtasks was constructed using three large-scale multilingual machine-generated text datasets --- HC3~\cite{guo-etal-2023-hc3}, M4GT-Bench~\cite{wang-etal-2024-m4gt}, and MAGE~\cite{li-etal-2024-mage}. We merged all collected data, removed repeated texts, and randomly split into train and development sets by the ratio of 7:3. See detailed distribution over different languages, domains and generators in \appref{app:traindevtest}.

\subsubsection{Dev-Test Set}

\textbf{English Subtask A:}
we utilized 13,371 human-written texts from RAID~\cite{dugan-etal-2024-raid} and sampled 19,186 MGTs from LLM-DetectAIve~\cite{mervat2024demo}. The latter contains MGTs of three types: (i) fully MGTs, (ii) human-written and then machine-polished texts, and (iii) machine-generated and then machine-humanized texts. 

\textbf{Multilingual Subtask B:}
we sampled data from MULTITuDE~\cite{macko-etal-2023-multitude} as the multilingual dev-test set.

\subsubsection{Test Set}
For the test set, in addition to leveraging MixSet~\cite{zhang-etal-2024-llm} and CUDRT~\cite{tao2024cudrt}, 
the majority of test sets is collected by our team, particularly multilingual subtask. Note that the dataset of CUDRT has not been released to the public before we used it as a subset of the test set. 

\textbf{English Subtask A} 
 uses Mixset and a subset of CUDRT. Based on the IELTS essays, we collected generations from \textit{Llama3.1-70B-versatile} and \textit{GPT-4o-mini}. We further generated academic paper peer reviews based on NLPeer and PeerSum, using \textit{GPT-4o} and \textit{GPT-4o-mini}. 

\textbf{Multilingual Subtask B:} in addition to two datasets --- we used Urdu fake news detection datasets generated by \citet{ali2024detection}, and sampled data from the CUDRT Chinese subset, the rest of multilingual test set was all newly collected, involving 27 different corpus and spanning 15 languages, with six of them are not seen in training, dev and dev-test sets.
It covers Arabic, Chinese, Dutch, German, \textbf{Hebrew}, \textbf{Hindi}, Indonesian, Italian, \textbf{Japanese}, \textbf{Kazakh}, \textbf{Norwegian}, Russian, Spanish, Urdu, and \textbf{Vietnamese} (languages highlighted with the bold font were not seen in the training data).\footnote{We included 15 languages in the training, dev and dev-test sets --- Arabic, \textbf{Bulgarian}, \textbf{Catalan}, Chinese, Czech, Dutch, \textbf{English}, \textbf{German}, Indonesian, Italian, \textbf{Portuguese}, Russian, Spanish, \textbf{Ukrainian}, and Urdu.}
See detailed distribution over sources, domains, and models in \appref{app:testset}.

\subsection{Baselines}

\paragraph{Detector}

We fine-tuned pre-trained transformer-based models on the training sets as baselines. For \textit{subtask A}, we fine-tuned RoBERTa, and XLM-R for \textit{subtask B} to handle with multilingual data.

Fine-tuning was performed using the Hugging Face \texttt{Trainer} API with the following configuration: learning rate of $2 \times 10^{-5}$, batch size of 16 for both training and evaluation, weight decay of 0.1, and a total of 3 training epochs. Models were evaluated at the end of each epoch, and we keep the best model determined by development set performance, for the subsequent testing.


\begin{table}[t]
\centering
\scriptsize
\resizebox{\columnwidth}{!}{

\begin{tabular}{lccc}
\toprule
\textbf{Task} & \textbf{Set} & \textbf{Accuracy} & \textbf{F1} \\
\midrule
\multirow{3}{*}{Subtask A} & Dev      & 96.2 & 95.9 / 96.2 \\
       & Dev-Test & 83.1 & 81.6 / 82.6 \\
       & Test     & 74.9 & 73.4 / 73.8 \\
\midrule
\multirow{3}{*}{Subtask B}& Dev      & 95.2 & 94.8 / 95.2 \\
       & Dev-Test & 84.7 & 65.5 / 85.7 \\
       & Test     & 74.7 & 74.2 / 74.3 \\
\bottomrule
\end{tabular}
}

\caption{Baseline performance on the Dev, Dev-Test, and Test sets for according to accuracy and macro F1.}
\label{tab:baseline_results}
\end{table}

\paragraph{Results on Dev, Dev-test, and Test Sets}

Baseline results on the dev, dev-test, and test sets for both subtask A and B are demonstrated in \tabref{tab:baseline_results}. The baseline models showed strong performance on the development (dev) sets, particularly for subtask A, achieving high accuracy and F1-scores. However, performance declined on the dev-test and test sets, indicating potential overfitting or challenges in adapting to unseen data distributions. 

For subtask B, the multilingual setting introduced additional complexity, as reflected in the relatively lower macro-average metrics, which emphasizes the difficulty of generalizing across multiple languages. These baseline results provide a reference point for participants and highlight the challenges of detecting machine-generated text, especially in multilingual contexts.

\section{Participants' Submissions}
In this section, we first describe ranking, macro-F1 and accuracy of participants, followed by a brief description of all submitted systems.
We classify methods into (1) above vs. below the baseline, (2) black-box vs. white-box, (3) zero-shot vs. fine-tuning, (4) fine-tuning based on small models vs. large models, and (5) ensemble or not. 

To describe systems participating in the English and Multilingual subtasks separately, in the text we add the subscript \textbf{English:rank} to participants in the English subtask and the subscript \textbf{Multi:rank} to participants in the multilingual subtask. For example the team \textbf{Fraunhofer SIT} is ranked 3rd in the English subtask and referred to as \textbf{Fraunhofer SIT$_{\text{English:3}}$} while it is ranked 10th in the Multilingual subtask and thus referred to as \textbf{Fraunhofer SIT$_{\text{Multi:10}}$}.

\subsection{English Subtask}

\subsubsection{Results and Rank}

\begin{table}[t]
    \centering
    \scriptsize
    \resizebox{\columnwidth}{!}{
    \begin{tabular}{llcc}
    \toprule
    \textbf{Rank} & \textbf{Team} & \textbf{Macro-F1} & \textbf{Accuracy} \\
    \midrule
    1 & Advacheck & 83.07 & 83.11 \\
    2 & Unibuc-NLP & 83.01 & 83.33 \\
    3 & Fraunhofer SIT & 82.80 & 82.89 \\
    4 & Grape  & 81.88 & 82.23 \\
    5 & TechExperts(IPN) & 81.53 & 81.81 \\
    6 & TurQUaz & 80.68 & 80.74 \\
    7 & SzegedAI & 79.10 & 79.29 \\
    8 & AAIG  & 78.74 & 79.34 \\
    9 & DCBU & 77.13 & 78.01 \\
    10 & Alfa & 75.37 & 76.42 \\
    11 & L3i++ & 74.63 & 75.54 \\
    12 & LuxVeri  & 74.58 & 75.68 \\
    13 & azlearning & 74.14 & 75.17 \\
    14 & honghanhh & 73.94 & 75.14 \\
    \hdashline
    -- & \textbf{Baseline} & 73.42 & 74.89 \\
    \hdashline
    15 & VX1291 & 72.93 & 74.83 \\ 
    -- & cuettransform & 72.32 & 73.16 \\
    16 & rockstart & 72.24 & 73.89 \\
    17 & batirsdu & 71.01 & 71.42 \\
    18 & IPN-CIC  & 70.68 & 72.42 \\
    19 & Ai-Monitors & 70.57 & 72.65 \\
    20 & semanticcuet & 70.05 & 71.96 \\
    21 & hmcgovern & 68.48 & 69.51 \\
    22 & abhirak0603 & 68.02 & 70.50 \\
    23 & cnlpnitspp & 65.02 & 68.76 \\
    24 & mail6djj & 64.66 & 68.46 \\
    25 & bennben & 63.32 & 67.48 \\
    26 & saehyunma & 62.80 & 67.25 \\
    27 & yuwert777 & 62.14 & 66.69 \\
    28 & seven & 59.09 & 63.20 \\
    29 & fangsifan & 58.48 & 62.68 \\
    30 & yaoxy & 57.28 & 64.20 \\
    31 & jojoc & 54.16 & 60.37 \\
    32 & dominikmacko & 49.94 & 50.78 \\
    33 & tropaleum & 49.57 & 50.60 \\
    34 & starlight1 & 47.57 & 56.65 \\
    35 & nitstejasrikar & 44.89 & 57.24 \\
    \bottomrule
    \end{tabular}
    }
    \caption{\textbf{English subtask} leaderboard results. The main performance metric is macro-F1. Accuracy is used as an auxiliary performance metric. 
    }
    \label{tab:results_en}
\end{table}

The English subtask attracted 36 submissions in total. \autoref{tab:results_en} presents the complete rankings. The competition saw a remarkably tight race among the top performers with only 0.27 macro-F1 points separating the top three teams: \textbf{Advacheck$_{\text{English:1}}$} (83.07), \textbf{Unibuc-NLP$_{\text{English:2}}$} (83.01), and \textbf{Fraunhofer SIT$_{\text{English:3}}$} (82.8). Interestingly, while the team \textbf{Advacheck$_{\text{English:1}}$} secured the first place by the main metric, \textbf{Unibuc-NLP$_{\text{English:2}}$} achieved a slightly higher accuracy (83.33 vs. 83.11), highlighting the razor-thin margins between top performers.

Fourteen teams outperformed the baseline (73.42 macro-F1) according to the main metric with scores varying from 83.07 to 44.89. The inability of most submissions to surpass the baseline underscores the complexity of the task.

\subsubsection{System Description}
\begin{table}[t]
    \resizebox{0.85\columnwidth}{!}{
    \centering
    \scriptsize
    \begin{tabular}{rccccc}
    \textbf{Team Name} & \rotatebox{90}{\textbf{Ranking}} & \rotatebox{90}{\textbf{Small PLM}} & \rotatebox{90}{\textbf{LLM}} & \rotatebox{90}{\textbf{Feature Combination}} & \rotatebox{90}{\textbf{Ensemble}} \\
       \toprule
       Advacheck & 1 & \checkmark & & & \\ 
       Unibuc-NLP & 2 & & \checkmark & & \\ 
       Fraunhofer SIT & 3 & \checkmark & & & \\ 
       Grape & 4 & \checkmark & \checkmark & & \checkmark \\ 
       TechExperts(IPN) & 5 & & \checkmark & & \\ 
       TurQUaz & 6 & \checkmark & \checkmark & \checkmark & \checkmark \\ 
       SzegedAI & 7 & \checkmark & & & \checkmark \\ 
       AAIG & 8 & \checkmark & & & \\ 
       DCBU & 9 & \checkmark & \checkmark & \checkmark & \checkmark \\ 
       L3i++ & 11 & & \checkmark & & \\
       LuxVeri & 12 & \checkmark & & \checkmark \\ 
       IPN-CIC & 18 & \checkmark & & & \\ 
       Ai-Monitors & 19 & \checkmark & & & \\ 
    \bottomrule
    \end{tabular}}
    \caption{\textbf{English subtask} participants overview.}
    \label{tab:systems_en}
\end{table}
\autoref{tab:systems_en} presents an overview of the English subtask participants' systems.\footnote{Top ranking teams that lack a system description do so because the authors did not submit a manuscripts and did not provide a short description of their system.}

\noindent \textbf{Team Advacheck$_{\text{English:1}}$} \cite{genai-detect:2025:task:Advacheck} develops a multi-task system with a shared Transformer Encoder (DeBERTa-v3-base) between several classification heads. This system includes a primary binary classification head and additional multiclass heads for text domain classification. The ablation studies show that multi-task learning outperforms single-task modes, with simultaneous tasks forming cluster structures in the embeddings space.

\noindent \textbf{Team Unibuc-NLP$_{\text{English:2}}$} \cite{genai-detect:2025:task:UnibucNLP} utilized both masked (XLM-RoBERTa-base) and causal language models (Qwen2.5-0.5B; \citet{yang2024qwen2}),\footnote{\url{https://qwenlm.github.io/blog/qwen2.5/ }} with the Qwen-based classifier performing on par with \citeauthor{genai-detect:2025:task:Advacheck}. The authors report that LORA fine-tuning XLM-RoBERTa promotes a strong performance.

\noindent \textbf{Team Fraunhofer SIT$_{\text{English:3}}$} \cite{genai-detect:2025:task:FraunhoferSIT} combined an MGT detection adapter with a multi-genre natural language inference adapter over RoBERTa-base. 

\noindent \textbf{Team Grape$_{\text{English:4}}$} \cite{genai-detect:2025:task:Grape}, first,
finetuned Llama-3.2-1B \cite{dubey2024llama} and gemma-2-2b \cite{team2024gemma} for the MGT detection task. Second, they combined linguistic features with the model predictions by leveraging ensemble learning for more robust classification.

\noindent \textbf{Team TechExperts(IPN)$_{\text{English:5}}$} 
similar to \citeauthor{genai-detect:2025:task:Grape} fine-tuned gemma-2b for the MGT detection task, which confirms the effectiveness of the small model in identifying the generated content.

\noindent \textbf{Other teams} ranked in top-20 developed the MGT detectors by (i) fine-tuning a model (\textbf{Team TurQUaz$_{\text{English:6}}$}; \citealp{genai-detect:2025:task:TurQUaz};  \textbf{Team AAIG$_{\text{English:8}}$}; \citealp{genai-detect:2025:task:AAIG}; \textbf{Team IPN-CIC$_{\text{English:18}}$}; \citealp{genai-detect:2025:task:abit7431}; \textbf{Team Ai-Monitors$_{\text{English:19}}$}; \citealp{genai-detect:2025:task:AI-Monitors}); (ii) ensembling models and features (\textbf{Team SzegedAI$_{\text{English:7}}$}; \citealp{genai-detect:2025:task:SzegedAI}; \textbf{Team DCBU$_{\text{English:9}}$}; \citealp{genai-detect:2025:task:dcbu}; \textbf{Team Lux Veri$_{\text{English:12}}$}; \citealp{genai-detect:2025:task:LuxVeri}); and (ii) utilizing label supervision (\textbf{Team L3i++$_{\text{English:11}}$}; \citealp{genai-detect:2025:task:L3i++}).

\subsection{Multilingual Subtask}

\subsubsection{Results and Ranks}

\begin{table}[t!]
    \centering
    \scriptsize
    \resizebox{\columnwidth}{!}{
    \begin{tabular}{llcc}
    \toprule
    \textbf{Rank} & \textbf{Team} & \textbf{Macro-F1} & \textbf{Accuray} \\
    \midrule
    1 & Grape & 79.16 & 79.62 \\
    -- & jykim* & 75.96 & 76.56 \\
    2 & rockstart & 75.57 & 75.64 \\
    3 & Nota AI & 75.32 & 75.91 \\
    4 & LuxVeri & 75.13 & 75.27 \\
    5 & TechExperts(IPN) & 74.63 & 74.74 \\
    6 & azlearning & 74.36 & 74.49 \\
    7 & nampfiev1995 & 74.27 & 74.40 \\
    \hdashline 
    -- & \textbf{Baseline} & 74.16 & 74.74 \\
    \hdashline
    8 & starlight1 & 73.78 & 73.92 \\
    9 & abit7431 & 72.65 & 73.48 \\
    10 & Fraunhofer SIT  & 72.58 & 73.61 \\
    11 & mail6djj & 72.24 & 73.34 \\
    12 & saehyunma & 72.20 & 73.52 \\
    13 & seven & 71.40 & 72.00 \\
    14 & jojoc & 70.72 & 70.99 \\
    15 & OSINT & 70.67 & 71.87 \\
    16 & yaoxy & 69.54 & 71.51 \\
    17 & VX1291 & 69.47 & 70.50 \\ 
    18 & bennben & 69.13 & 69.63 \\
    19 & fangsifan & 68.60 & 69.57 \\
    20 & yuwert777 & 68.45 & 70.65 \\
    21 & honghanhh & 67.61 & 67.91 \\
    22 & tmarchitan & 66.29 & 67.11 \\
    -- & keles & 64.24 & 64.41 \\
    23 & batirsdu & 62.59 & 63.05 \\
    24 & sohailwaleed2 & 52.53 & 52.59 \\
    25 & dominikmacko & 51.03 & 51.05 \\
    \bottomrule
    \end{tabular}
    }
    \caption{\textbf{Multilingual subtask} leaderboard results. Submissions marked with ``*'' use additional training data and, therefore, are not incorporated in the ranking. 
    }
    \label{tab:results_multi}
\end{table}

The multilingual subtask received 27 submissions with complete rankings demonstrated in \autoref{tab:results_multi}.

The most notable feature of this subtask was the exceptional performance of the team ``Grape'', achieving macro-F1 score of 79.16, significantly outperforming other competitors. A substantial gap of 3.59 macro-F1 points between the winner and the second place ``rockstart'' (75.57) underscores the effectiveness of the ``Grape'' team approach to multilingual MGT detection.

In this subtask, only seven teams managed to surpass the baseline score of 74.16 with scores ranging from 79.16 to 51.03. This indicates the increased difficulty of detecting MGT text among multiple languages simultaneously.

The overall lower scores in this subtask compared to the English subtask (top score 79.16 vs. 83.07) highlight the additional complexity introduced by multilingual detection and room for improvement. 

\subsubsection{System Description}
\begin{table}[t!]
\centering
\resizebox{0.8\columnwidth}{!}{
    \scriptsize
    \begin{tabular}{rccccc}
    \textbf{Team Name} & \rotatebox{90}{\textbf{Ranking}} & \rotatebox{90}{\textbf{Small PLM}} & \rotatebox{90}{\textbf{LLM}} & \rotatebox{90}{\textbf{Feature Combination}} & \rotatebox{90}{\textbf{Ensemble}} \\
       \toprule
       Grape & 1 & \checkmark & \checkmark & & \checkmark \\ 
       Nota AI & 3 & \checkmark & \checkmark & \checkmark & \checkmark \\ 
       LuxVeri & 4 & \checkmark & & & \checkmark \\ 
       TechExperts(IPN) & 5 & & \checkmark & & \\ 
       Fraunhofer SIT & 10 & \checkmark & & & \\ 
       OSINT & 15 & \checkmark & & & \\ 
    \bottomrule
    \end{tabular}}
    \caption{\textbf{Multilingual subtask} participants overview.}
    \label{tab:systems_multi}
\end{table}

\autoref{tab:systems_multi} presents an overview of the multilingual subtask participants' systems. 

\noindent \textbf{Team Grape$_{\text{Multi:1}}$}~\citep{genai-detect:2025:task:Grape}, ranked 1 in the multilingual leaderboard, adopted two approaches in the task. They first separately fine-tuned small language models tailored to the specific subtask and then trained an ensemble model on top of them. Through evaluating and comparing these approaches, the team identified the most effective techniques for detecting machine-generated content across languages.


\noindent \textbf{Team NotaAI$_{\text{Multi:3}}$}~\citep{genai-detect:2025:task:NotaAI} secured the third place in the task. They developed the system that addresses the challenge of detecting MGT in languages not observed during training, where model accuracy tends to decline significantly. The proposed multilingual MGT detection system employs a two-step approach: first, a language identification tool determines the language of the input text. If the language has been observed during training, the text is processed using a model fine-tuned on a multilingual PLM. For languages not seen during training, the system utilizes a model that combines token-level predictive distributions extracted from various LLMs with a meaning representation derived from a multilingual PLM.

\noindent \textbf{Team LuxVeri$_{\text{Multi:4}}$}~\citep{genai-detect:2025:task:LuxVeri} earned the 4th place. They utilized an ensemble of models, where weights are assigned based on each model's inverse perplexity to improve classification accuracy. The system combined RemBERT, XLM-RoBERTa-base, and BERT-base-multilingual-cased using the same weighted ensemble strategy. The results highlight the effectiveness of inverse perplexity-based weighting for robust detection of machine-generated text in both monolingual and multilingual settings.

\noindent \textbf{Team TecExperts(IPN)$_{\text{Multi:5}}$}~\citep{genai-detect:2025:task:TecExperts(IPN)} leveraged the gemma-2b model, fine-tuned specifically for the Shared Task 1 datasets to achieve strong performance.


\noindent \textbf{Team L3i++$_{\text{Multi:7}}$}~\citep{genai-detect:2025:task:L3i++} studied a label-supervised adaptation configuration for LLaMA-as-a-judge for the task. In detail, they explore the feasibility of fine-tuning LLaMA with label supervision in masked and unmasked, unidirectional and bidirectional settings, to discriminate the texts generated by machines and humans in monolingual and multilingual corpora.

\noindent \textbf{Other Systems} The other systems explored various approaches, including exploring the integration of additional features such as perplexity and Tf-IDF~(\textbf{Team TurQUaz$_{\text{Multi:22}}$};~\citealp{genai-detect:2025:task:TurQUaz}), finetuning models such as XLM-RoBERTa on the training set for the final evaluation, as incorporating adapter fusion led to worse results~(\textbf{Team Fraunhofer SIT$_{\text{Multi:10}}$};~\citealp{genai-detect:2025:task:FraunhoferSIT}), XML-R and mBERT models~(\textbf{Team IPN$_{\text{Multi:9}}$};~\citealp{genai-detect:2025:task:abit7431} and QWen and RoBERTa models~(\textbf{Team Unibuc-NLP$_{\text{Multi:22}}$};~\citealp{genai-detect:2025:task:UnibucNLP}); and combining language-specific embeddings with fusion techniques to create a unified, language-agnostic feature representation~(\textbf{Team OSINT$_{\text{Multi:15}}$};~\citealp{genai-detect:2025:task:OSINT}).





\begin{table}[t!]
    \centering
    \small
    \resizebox{\columnwidth}{!}{
    \begin{tabular}{cccccc}
    \toprule
    \textbf{Rank} & \textbf{All} & \textbf{MixSet} & \textbf{CUDRT} & \textbf{IELTS} & \textbf{PeerReview} \\
    \midrule
    1 & 83.1 & 48.0 & 67.1 & 89.9 & 97.2 \\
    2 & 83.3 & 66.7 & 75.9 & 82.6 & 94.1 \\
    3 & 82.9 & 58.9 & 71.0 & 88.8 & 92.1 \\
    4 & 82.2 & 64.7 & 73.2 & 79.1 & 97.4 \\
    5 & 81.8 & 59.2 & 72.7 & 80.8 & 95.5 \\
    6 & 80.7 & 47.2 & 72.6 & 78.1 & 96.9 \\
    7 & 75.7 & 54.9 & 71.0 & 63.1 & 97.2 \\
    8 & 79.3 & 62.3 & 75.4 & 69.0 & 97.2 \\
    9 & 78.0 & 60.0 & 74.6 & 66.3 & 96.9 \\
    10 & 76.4 & 59.8 & 75.5 & 64.2 & 93.2 \\
    11 & 75.5 & 60.9 & 70.3 & 66.9 & 92.5 \\
    12 & 75.7 & 56.6 & 74.0 & 61.9 & 95.2 \\
    13 & 75.2 & 62.8 & 70.8 & 65.3 & 92.2 \\
    14 & 75.1 & 66.6 & 72.8 & 62.7 & 92.2 \\
    \hdashline 
    \textbf{BL} & 74.9 & 62.0 & 72.1 & 63.4 & 92.2 \\
    \hdashline 
    15 & 74.8 & 73.2 & 71.9 & 63.0 & 90.8 \\
    - & 73.2 & 53.5 & 71.3 & 62.8 & 89.3 \\
    16 & 73.9 & 64.3 & 71.2 & 62.6 & 90.3 \\
    17 & 71.4 & 53.9 & 69.6 & 70.8 & 76.6 \\
    18 & 72.4 & 65.4 & 70.6 & 62.2 & 86.5 \\
    19 & 72.7 & 72.6 & 70.4 & 63.6 & 84.8 \\
    20 & 72.0 & 69.8 & 70.4 & 66.5 & 79.8 \\
    21 & 69.5 & 50.7 & 64.0 & 65.7 & 82.0 \\
    22 & 70.5 & 70.6 & 66.7 & 65.3 & 80.0 \\
    23 & 68.8 & 73.7 & 66.9 & 61.7 & 77.6 \\
    24 & 68.5 & 65.7 & 67.3 & 57.4 & 82.0 \\
    25 & 67.5 & 67.6 & 67.7 & 58.0 & 77.5 \\
    26 & 67.2 & 68.2 & 67.2 & 57.3 & 78.0 \\
    27 & 66.7 & 67.4 & 67.1 & 57.1 & 76.5 \\
    28 & 63.2 & 68.3 & 67.8 & 57.1 & 64.4 \\
    29 & 63.5 & 67.7 & 68.6 & 57.6 & 64.0 \\
    30 & 64.2 & 77.7 & 64.5 & 58.6 & 67.9 \\
    31 & 60.4 & 77.7 & 64.6 & 58.3 & 55.6 \\
    32 & 50.8 & 56.0 & 49.7 & 51.1 & 50.7 \\
    33 & 50.6 & 56.7 & 49.1 & 50.7 & 51.0 \\
    34 & 56.6 & 80.8 & 60.6 & 54.9 & 50.9 \\
    35 & 57.2 & 82.3 & 56.4 & 54.0 & 57.8 \\
    \bottomrule
    \end{tabular}
}
    \caption{\textbf{English subtask detection accuracy} across four domains.}
    \label{tab:ana-english-domain-impact}
\end{table}

\begin{table*}[t]
    \centering
    \small
    \begin{tabular}{cccccccccc}
    \toprule
    \textbf{Rank} & \textbf{All} & \textbf{News} & \textbf{Wiki} & \textbf{Essay} & \textbf{QA} & \textbf{Summary} & \textbf{Tweet} & \textbf{GovR} &\textbf{ Other} \\
    Size & 151,425 & 57,590 & 11,687 & 2,201 & 24,854 & 13,600 & 1,325 & 19,736 & 4,214 \\
    \midrule
    1 & 79.6 & 65.1 & 80.2 & 99.3 & 98.9 & 70.0 & 94.5 & 87.0 & 84.2 \\
    2 & 75.6 & 64.0 & 87.1 & 81.0 & 91.9 & 79.1 & 100.0 & 69.1 & 48.2 \\
    3 & 75.9 & 60.7 & 81.0 & 97.7 & 96.2 & 65.2 & 72.0 & 81.7 & 91.1 \\
    4 & 75.3 & 60.7 & 87.9 & 91.0 & 93.2 & 71.7 & 98.9 & 75.2 & 58.6 \\
    \hdashline
        \textbf{BL} & 74.8 & 61.6 & 85.2 & 97.7 & 94.1 & 58.6 & 94.4 & 76.2 & 83.2 \\
    \hdashline
    5 & 74.7 & 60.2 & 74.7 & 97.7 & 98.9 & 59.7 & 65.3 & 75.0 & 96.2 \\
    6 & 74.5 & 59.8 & 79.6 & 90.9 & 95.1 & 82.8 & 95.5 & 62.6 & 82.7 \\
    7 & 74.4 & 59.8 & 79.7 & 90.7 & 95.2 & 82.1 & 93.8 & 62.9 & 79.4 \\
    8 & 73.9 & 58.1 & 81.2 & 98.5 & 92.9 & 73.5 & 29.1 & 81.2 & 70.7 \\
    9 & 73.5 & 61.1 & 85.0 & 94.7 & 94.5 & 64.8 & 87.8 & 78.7 & 60.3 \\
    10 & 73.6 & 60.8 & 77.3 & 94.2 & 95.4 & 61.3 & 91.9 & 80.5 & 86.8 \\
    11 & 73.3 & 60.2 & 83.9 & 96.7 & 94.9 & 60.0 & 56.0 & 82.4 & 61.8 \\
    12 & 73.5 & 62.2 & 81.4 & 93.3 & 95.9 & 64.8 & 41.0 & 83.5 & 68.2 \\
    13 & 72.0 & 56.3 & 42.3 & 99.2 & 99.2 & 70.9 & 33.7 & 89.0 & 67.3 \\
    14 & 71.0 & 56.0 & 55.2 & 97.0 & 92.4 & 76.3 & 0.1 & 81.1 & 85.6 \\
    15 & 50.3 & 51.0 & 42.4 & 60.0 & 51.2 & 49.7 & 33.9 & 61.9 & 62.1 \\
    16 & 71.5 & 59.6 & 44.0 & 97.0 & 99.2 & 59.5 & 57.7 & 89.3 & 58.1 \\
    17 & 50.2 & 50.8 & 43.2 & 57.7 & 50.7 & 49.9 & 36.6 & 59.8 & 60.8 \\
    18 & 69.6 & 55.0 & 45.8 & 97.7 & 92.2 & 71.5 & 2.3 & 82.7 & 85.2 \\
    19 & 70.5 & 54.5 & 33.5 & 99.1 & 99.1 & 73.1 & 6.4 & 88.7 & 77.6 \\
    20 & 70.7 & 60.9 & 41.7 & 93.5 & 99.1 & 63.5 & 45.3 & 86.8 & 61.3 \\
    21 & 67.9 & 61.7 & 69.9 & 63.6 & 78.1 & 78.0 & 49.4 & 71.8 & 60.7 \\
    22 & 67.1 & 57.4 & 51.8 & 83.4 & 94.7 & 61.5 & 100.0 & 80.7 & 20.9 \\
    23 & 49.7 & 49.1 & 57.0 & 45.5 & 49.1 & 50.3 & 64.5 & 40.1 & 39.4 \\
    24 & 52.6 & 45.3 & 35.0 & 83.0 & 72.4 & 67.3 & 99.3 & 46.6 & 17.8 \\
    25 & 51.0 & 50.4 & 53.0 & 51.0 & 51.8 & 52.0 & 56.1 & 48.4 & 48.9 \\
    \bottomrule
    \end{tabular}
    \caption{\textbf{Multilingual subtask detection accuracy} across eight domains (Wiki: Wikipedia, GovR: GovReport).}
    \label{tab:ana-multilingual-domain-impact}
\end{table*}


\section{Analysis}
Based on the test set, we analyze submitted systems by comparing the detection accuracy on (1) in-domain vs. out-of-domain, (2) seen vs. unseen languages, and (3) generations produced using normal prompts vs. prompts attempting to fill the gap between human and machine based on observations in manual annotations.

\subsection{English In-domain vs. Out-of-domain}
Results in \tabref{tab:ana-english-domain-impact} show the accuracy of 36 submitted systems across four component datasets in the English test set. Significant variance across domains reveals different generalization and robustness across detection systems.

Performance for in-domain datasets, such as IELTS and PeerReview, is generally higher than out-of-domain datasets MixSet and CUDRT. Top systems ranking 1-5 achieve scores around 80\% on in-domain datasets. For example, top1 Team ``Advacheck'' scored 83.1\% on IELTS essays and 89.9\% on PeerReview. Moreover, accuracies are $\geq$90\% for all teams above the baseline on PeerReview including the baseline itself. 
The consistently-high performance suggests that peer reviews (PeerRead) in the M4GT-Bench training set have effectively facilitated detectors in capturing domain-specific patterns during training, and thus generalizing well to similar-content PeerReview in the test set. 
For IELTS essays, the performance trend differs slightly from PeerReview. Despite student essays presented in the training set M4GT-Bench, only the first five teams managed to achieve scores $\geq$80\%. This lies in the fact that essays sampled from OUTFOX in M4GT-Bench were written by English native speakers, while English is the second language for authors who attended the IELTS test. 
Subtle differences between essays in the training and test result in accuracy declines on the test set, which to some extent reveals the vulnerability of detectors against tiny distribution perturbations. 

Out-of-domain dataset MixSet is the most challenging subset due to its varied and unseen content genres including game reviews, email, blog, and speech content. 
Top-ranked teams (ranks 1–5) 
experienced a substantial performance drop on MixSet --- accuracy in the range of 48–66.7\%. This may also attribute to the humanization and adaption of machine-generated text in MixSet. The former refers to modifying MGT to more closely mimic the natural noise that human writing always brings, introducing typo, grammatical mistakes, links, and tags. The latter refers to modifying MGT to ensure its alignment to fluency and naturalness to human linguistic habits without introducing any error expression.
Detection systems struggle with highly heterogeneous and less structured data, which is exacerbated by the humanization and adaption operations of MGT in MixSet.

A surprising observation on MixSet is that all teams above the baseline struggled to improve $\leq$5\% compared to the baseline 62\%, while 
15 teams below the baseline achieved improvements $\geq$5\%, with remarkable scores achieved by the last two teams --- 80.8\% and 82.3\%, showing a stark contrast to their performance on other datasets.

Domains involved in CUDRT partially overlapped with the training data domains (e.g., news), while thesis is out of the training data though similar to academic papers, leading to the accuracy between Mixset and PeerReview. Most teams including the baseline scored between 65–75\%, demonstrating moderate adaptability to this dataset.

\subsection{Multilingual Subtask}
We analyze submissions from three perspectives.

\subsubsection{In-domain vs. Out-of-domain}
We divided 29 sources across 15 languages into 8 domains: News, Wikipedia, Essay, question answering (QA), Summary, Tweet, government reports (GovReport), and others (e.g., poetry).

\autoref{tab:ana-multilingual-domain-impact} presents the multilingual Subtask accuracy across 8 domains. In-domain datasets (News, Wikipedia, QA and Summary) consistently achieve higher accuracies due to their structured and training-aligned nature. Baseline accuracies for these domains are relatively strong, with significant improvements by the top-performing teams. Notably, the top-ranked team achieved peak performance of over 98\% in QA, while the second-ranked team attained over 87\% in Wikipedia.
Though the genre of summary presented in the training data, they are English text. Summaries in the test set are Russian and Arabic, so summary domain posed notable challenges for detector, performing poorly across both baselines and team submissions. This underscores the difficulty of distinguishing machine-generated summaries from human-written ones in this domain.
 
Conversely, out-of-domain datasets (Essay, Tweet, GovReport, and Other) presented greater challenges, reflecting the systems' struggles to generalize to unseen styles or informal text. While structured datasets like essays and GovReport performed moderately well, with top-team accuracies exceeding 85\%, informal and noisy domains such as tweets exhibited the lowest performance, with accuracies peaking at just 69.99\%. This stark contrast highlights the need for more effective generalization strategies. Interestingly, we observed an anomaly in the tweet domain: two teams (ranked second and 22nd) achieved perfect accuracy (100\%). This suggests that specialized approaches tailored to this domain can yield exceptional results, though these may involve overfitting to specific dataset patterns.

Overall, the results reveal a persistent gap between in-domain and out-of-domain performance, emphasizing the importance of domain adaptation and robust methods for handling unstructured or unseen data. At the same time, the findings demonstrate the potential for domain-specific optimizations in challenging contexts.

\begin{table}[t!]
    \centering
    \small
    \resizebox{\columnwidth}{!}{
    \begin{tabular}{ccccc}
    \toprule
    \textbf{Rank} & \textbf{All} & \textbf{Fill-gap} & \textbf{Original} & \textbf{Others}\\
    Size & 151,425 & 32,487 & 17,017 &  101,921 \\
    \midrule
    1 & 79.6 & 91.1 & 94.2 & 73.5 \\
    2 & 75.6 & 75.9 & 84.0 & 74.1 \\
    3 & 75.9 & 89.7 & 92.2 & 68.8 \\
    4 & 75.3 & 81.5 & 86.9 & 71.4 \\
    \hdashline 
    \textbf{BL} & 74.8 & 87.6 & 89.0 & 68.3 \\
    \hdashline
    5 & 74.7 & 84.6 & 96.6 & 67.9 \\
    6 & 74.5 & 75.6 & 90.1 & 71.5 \\
    7 & 74.4 & 75.4 & 90.3 & 71.4 \\
    8 & 73.9 & 88.5 & 87.1 & 67.0 \\
    9 & 73.5 & 86.7 & 93.1 & 66.0 \\
    10 & 73.6 & 92.9 & 93.0 & 64.2 \\
    11 & 73.3 & 88.3 & 91.6 & 65.5 \\
    12 & 73.5 & 91.6 & 94.3 & 64.3 \\
    13 & 72.0 & 93.7 & 95.7 & 61.1 \\
    14 & 71.0 & 90.4 & 86.3 & 62.3 \\
    15 & 50.3 & 66.7 & 64.8 & 42.7 \\
    16 & 71.5 & 93.2 & 96.4 & 60.4 \\
    17 & 50.2 & 64.7 & 62.9 & 43.5 \\
    18 & 69.6 & 91.6 & 86.5 & 59.8 \\
    19 & 70.5 & 94.9 & 95.1 & 58.6 \\
    20 & 70.7 & 93.8 & 96.1 & 59.0 \\
    21 & 67.9 & 79.9 & 71.5 & 63.5 \\
    22 & 67.1 & 84.6 & 94.4 & 57.0 \\
    23 & 49.7 & 36.1 & 37.4 & 56.1 \\
    24 & 52.6 & 66.4 & 60.3 & 46.9 \\
    25 & 51.0 & 48.2 & 48.5 & 52.4 \\
    \bottomrule
    \end{tabular}
}
    \caption{\textbf{Multilingual subtask detection accuracy} between generations using original prompts vs. prompts aiming to fill the gap between human and machine, corresponding to columns of \textit{Original} vs. \textit{Fill-gap}. All is the whole multilingual test set.}
    \label{tab:ana-prompt-impact}
\end{table}

\begin{table*}[t!]
    \centering
    \small
    \resizebox{\textwidth}{!}{
    \begin{tabular}{ccccccccccccccccc}
    \toprule
    \textbf{Rank} & \textbf{All} & \textbf{ZH} & \textbf{UR} & \textbf{RU} & \textbf{AR} & \textbf{IT} & \underline{\textbf{KK}} & \underline{\textbf{VI}} & \textbf{DE} & \underline{\textbf{NO}} & \textbf{ID} & \textbf{NL} & \textbf{ES} & \underline{\textbf{HI}} & \underline{\textbf{HE}} & \underline{\textbf{JA}} \\
    Size & 151,425 & 63,009 & 30,505 & 27,158 & 10,670 & 5,296 & 2,471 & 2,326 & 1,865 & 1,544 & 1,200 & 1,200 & 1,200 & 1,199 & 1,182 & 600 \\
    \midrule
    1 & 79.6 & 94.2 & 68.7 & 67.1 & 71.2 & 52.9 & 55.5 & 90.5 & 88.3 & 80.3 & 89.6 & 82.2 & 89.5 & 51.8 & 86.7 & 77.0 \\
    2 & 75.6 & 84.7 & 64.6 & 74.2 & 57.9 & 52.9 & 83.8 & 83.5 & 96.4 & 76.0 & 51.7 & 90.6 & 91.2 & 69.6 & 96.8 & 95.3 \\
    3 & 75.9 & 90.2 & 67.2 & 58.9 & 66.8 & 52.9 & 92.5 & 74.7 & 88.8 & 72.2 & 87.4 & 68.9 & 47.1 & 70.6 & 96.4 & 72.2 \\
    4 & 75.3 & 87.6 & 64.6 & 63.9 & 61.3 & 52.9 & 75.8 & 83.4 & 94.9 & 88.5 & 53.5 & 92.2 & 90.4 & 73.0 & 97.3 & 92.2 \\
    \hdashline 
    \textbf{BL} & 74.8 & 87.3 & 68.4 & 55.3 & 68.4 & 52.9 & 82.8 & 85.3 & 85.2 & 69.8 & 68.2 & 92.5 & 90.5 & 71.3 & 89.3 & 90.0 \\
    \hdashline
    5 & 74.7 & 90.1 & 64.1 & 56.0 & 69.1 & 52.9 & 62.9 & 87.6 & 59.6 & 69.8 & 93.8 & 81.0 & 90.4 & 69.1 & 96.5 & 95.0 \\
    6 & 74.5 & 84.2 & 65.0 & 67.9 & 66.8 & 52.9 & 47.5 & 81.8 & 93.5 & 83.2 & 83.9 & 85.9 & 88.9 & 69.1 & 89.8 & 78.2 \\
    7 & 74.4 & 84.4 & 64.9 & 67.7 & 65.4 & 52.9 & 47.5 & 82.0 & 92.2 & 85.8 & 83.4 & 85.4 & 89.2 & 68.8 & 90.1 & 75.2 \\
    8 & 73.9 & 88.3 & 58.7 & 67.0 & 58.4 & 52.9 & 93.0 & 65.9 & 89.6 & 61.6 & 50.5 & 80.7 & 88.0 & 61.4 & 82.7 & 61.2 \\
    9 & 73.5 & 85.1 & 67.0 & 59.8 & 60.8 & 52.9 & 90.6 & 87.2 & 82.8 & 78.2 & 48.7 & 78.0 & 83.1 & 54.5 & 89.6 & 74.3 \\
    10 & 73.6 & 86.0 & 67.6 & 56.0 & 69.1 & 52.9 & 86.8 & 80.4 & 65.0 & 52.8 & 73.8 & 87.4 & 85.4 & 63.5 & 85.7 & 86.0 \\
    11 & 73.3 & 87.4 & 63.4 & 58.2 & 55.6 & 52.9 & 89.4 & 79.7 & 87.0 & 66.6 & 73.9 & 82.1 & 87.4 & 70.5 & 93.3 & 79.5 \\
    12 & 73.5 & 85.3 & 68.0 & 61.5 & 54.3 & 52.9 & 92.7 & 62.0 & 87.8 & 63.7 & 80.3 & 85.3 & 86.3 & 63.0 & 86.2 & 59.5 \\
    13 & 72.0 & 93.2 & 55.4 & 63.3 & 55.4 & 52.9 & 93.0 & 65.9 & 5.2 & 25.8 & 71.2 & 50.2 & 50.0 & 61.4 & 1.7 & 61.2 \\
    14 & 71.0 & 87.0 & 54.3 & 68.7 & 61.2 & 52.8 & 54.7 & 63.8 & 77.1 & 54.7 & 49.7 & 57.1 & 64.9 & 53.5 & 0.0 & 52.0 \\
    15 & 50.3 & 50.9 & 52.0 & 49.0 & 53.0 & 50.4 & 52.1 & 49.7 & 33.9 & 33.2 & 49.7 & 50.3 & 50.7 & 50.4 & 32.1 & 50.0 \\
    16 & 71.5 & 91.3 & 62.4 & 55.5 & 53.7 & 52.9 & 89.4 & 79.7 & 5.3 & 28.9 & 79.9 & 50.2 & 50.0 & 70.3 & 1.9 & 79.5 \\
    17 & 50.2 & 50.6 & 51.4 & 49.3 & 52.8 & 50.1 & 52.2 & 50.1 & 35.9 & 34.5 & 49.3 & 50.3 & 50.2 & 50.6 & 34.2 & 53.3 \\
    18 & 69.6 & 87.4 & 54.5 & 63.8 & 61.1 & 52.9 & 55.7 & 57.0 & 58.2 & 23.1 & 50.3 & 55.2 & 59.3 & 53.7 & 0.0 & 54.3 \\
    19 & 70.5 & 92.2 & 51.6 & 65.5 & 56.5 & 52.8 & 54.7 & 63.8 & 4.2 & 23.8 & 70.6 & 50.1 & 50.0 & 53.5 & 0.0 & 52.0 \\
    20 & 70.7 & 87.6 & 65.6 & 58.3 & 52.0 & 52.9 & 92.7 & 62.0 & 5.0 & 28.2 & 81.7 & 50.2 & 50.0 & 63.0 & 1.9 & 59.5 \\
    21 & 67.9 & 71.9 & 51.7 & 80.1 & 55.3 & 78.3 & 48.1 & 63.8 & 93.8 & 82.1 & 72.4 & 83.5 & 84.7 & 52.3 & 31.7 & 63.8 \\
    22 & 67.1 & 82.5 & 61.5 & 55.3 & 45.8 & 52.9 & 94.2 & 71.6 & 12.0 & 27.9 & 57.5 & 63.3 & 73.6 & 53.5 & 20.3 & 57.2 \\
    23 & 49.7 & 49.2 & 48.4 & 50.7 & 47.4 & 49.0 & 50.3 & 49.7 & 65.5 & 63.5 & 50.4 & 51.1 & 49.2 & 51.9 & 64.5 & 52.0 \\
    24 & 52.6 & 60.7 & 45.7 & 58.9 & 28.8 & 52.9 & 47.5 & 48.1 & 5.8 & 39.8 & 47.7 & 49.5 & 51.2 & 46.0 & 5.8 & 27.0 \\
    25 & 51.0 & 51.1 & 49.9 & 51.5 & 50.8 & 50.1 & 50.1 & 52.3 & 55.9 & 54.5 & 52.5 & 54.0 & 49.9 & 52.4 & 53.7 & 52.0 \\
    \bottomrule
    \end{tabular}
    
}
    \caption{\textbf{Multilingual subtask detection accuracy} across 15 languages. \underline{Underlined languages} were not present in the training data.}
    \label{tab:ana-language-impact}
\end{table*}

\subsubsection{General Prompts vs. Improved Prompts}

We compare system's accuracy results on text generated by ordinary prompts and the well-designed prompts that are used to fill the human and machine generations gap. MGTs using the improved prompts appear to make detection tasks more challenging.
Our improved prompts aim to make machine-generated text more similar to human-written text by instructing LLMs how to generate human-like text and to avoid presenting distinguishable signals in formats, where these features were summarized from our observations in manual annotations in distinguishing human and machine text. 

As shown in \tabref{tab:ana-prompt-impact}, in scenarios where detectors are tasked with identifying machine-generated text created using our improved prompts (Fill-gap in the \tabref{tab:ana-prompt-impact}), there is a noticeable decrease in accuracy compared to detecting machine-generated text created with the original prompts. This decline is particularly evident in higher ranks, with team 2 experiencing an 8\% drop, team 5 a 12\% drop, and teams 6 and 7 around a 15\% drop.
This decrease in performance suggests that the improved prompts, which were designed to narrow the gap between machine-generated and human-generated texts, may have inadvertently made the machine output too similar to human-like text, complicating the detector's ability to distinguish between the two.
However, there are exceptions to this trend. Notably, team 8 (rank 8) and team 14 (rank 14) show improved results when using Fill-gap prompts, with accuracy increasing from 87.08\% to 88.55\% for team 8 and from 86.30\% to 90.39\% for team 14. This improvement may be due to a misalignment of features between their detector design and our improved machine-generated prompt design.

This suggests that we can learn from machine-generated examples to design better prompts that make the machine-generated text more natural and less detectable. However, it also exposes the vulnerability of detectors --- they can be easily fooled when we adjust the prompts.

\subsubsection{Seen Languages vs. Unseen Languages}

\autoref{tab:ana-language-impact} presents the detection accuracy on the multilingual subtask across 15 languages, including seen and unseen languages during the training process.
The top-performing languages in terms of detection accuracy are generally those seen during training, with 
the highest accuracy observed on Chinese (94.2), followed by Russian (89.6) and Spanish (89.5). For Arabic (AR), Italian (IT), and Dutch (NL), the performance is slightly lower but still competitive, demonstrating the model’s steady generalization to seen languages.

For unseen languages, such as Hindi (HI) and Hebrew (HE), there is a noticeable drop in performance compared to seen languages. For example, the top-performing team achieved only 51.8 on Hindi. It is challenging for models to generalize to unseen languages, due to the limited exposure to linguistic patterns, structures, and features during training.
It is worth noting that some unseen languages perform relatively well, such as Kazakh (KK) and Vietnamese (VI), achieving relatively high scores. This may result from knowledge transfer from similar languages to the unseen, like Russian to Kazakh, and Chinese to Vietnamese.

Overall, the models perform well on seen languages, and scores decline significantly on unseen languages.The dataset size and the nature of a language (e.g., script, structure, and linguistic features) play an important role in the model’s ability to generalize.

\section{Conclusion}
In this work, we presented the dataset, baseline, participating systems and a detailed analysis across various detection methods for GenAI shared task 1: binary machine generated text detection. We explored both English and multilingual settings with diverse domains, LLM generators, and languages. All submitted systems show good performance on domains and languages that are seen during training, while witness the significant declines on unseen domains and languages. Moreover, detectors show remarkable vulnerability when machine-generated text is adapted to mimic humans, either by introducing typo, link, and tags, or by using fill-human-machine gap prompts. 
We expect our task can attract more researchers to develop robust and generalized detection models, and our analysis insights can provide a direction for future work, advancing research in machine-generated content detection. 


\section*{Limitations}
Despite providing a comprehensive dataset that spans multiple generators and domains and testing both English and Multilingual settings our study encounters several limitations that pave the way for future research.

Firstly, all the text samples (human and machine generated) used in this work come from existing open-source datasets and resources. While the sources of the test set have not been released prior to the conclusion of the challenge there is a limited possibility of data leakage. Participants were not allowed to use any external data and we trust they did not, however, pre-trained models could have seen part of the test set during their training and it would be impossible to know it.

Secondly, we don't have a detailed analysis of the differences between the datasets we joined together so that it is hard to understand if they have replicated or near-replicated samples and more in general how similar or not they are. In the future we will try to measure the performance of MGT detectors trained on the train set of one of these datasets when tested on each of the others to measure how close are the distributions of each pair of datasets among those we used. 

Finally, we only look at binary classification tasks (human vs. machine) while it would be relevant to understand the performance of detectors in a multiclass classification scenario (human vs. machine1 vs. machine2 vs. ...), this would have been difficult to arrange correctly using the different datasets we have collected since isolating the specific versions of each model becomes harder over time (specifically with closed source ones) and therefore we avoided doing it. Future work should account for this scenario too.



\section*{Ethics and Broader Impact}
This section outlines potential ethical considerations related to our work.

\paragraph{Data Collection and Licenses}
A primary ethical consideration is the data license. We reused pre-existing dataset, such as HC3, M4GT-Bench, MAGE, RAID, OUTFOX and LLM-DetectAIve, which have been publicly released for research purposes under clear licensing agreements. We adhere to the intended usage of all these dataset licenses.

\paragraph{Security Implications}
The dataset underpinning our shared task aims to foster the development of robust MGT detection systems, which are vital in addressing security and ethical concerns. These systems play a crucial role in identifying and mitigating misuse cases, such as preventing the spread of automated misinformation campaigns, which can undermine public discourse, and protecting individuals and organizations from potential financial losses through deceptive machine-generated content. In sensitive domains like journalism, academia, and legal proceedings, where the authenticity and accuracy of information are incredibly important, MGT detection is vital to maintaining content integrity and public trust. Beyond these fields, robust detection mechanisms contribute to the broader goal of promoting digital literacy by raising public awareness of the strengths and limitations of LLMs. This fosters a healthy skepticism towards digital content, encouraging users to critically evaluate the information they encounter.

Moreover, in multilingual contexts, detecting MGT becomes significantly more challenging due to the diversity of linguistic and cultural nuances. Advanced detection systems should address these complexities to prevent vulnerabilities, such as exploitation of less-resourced languages for disinformation. By ensuring the reliability of multilingual machine-generated content, these systems enhance global trust in AI technologies and protect against the security risks that arise from their misuse.


\bibliography{ref}
\bibliographystyle{acl_natbib}

\clearpage
\onecolumn
\section*{Appendix}
\appendix

\section{Dataset Distributions}
\label{app:dataset-dist}

\subsection{Training and Development Sets}
\label{app:traindevtest}
Tables \ref{tab:english-traindev} and \ref{tab:multi-traindev} respectively demonstrate the statistical information of training and development sets across different sources of English and multilingual subtasls, and \tabref{tab:generator-traindev-format} shows the distribution over generators for datasets of HC3, M4GT-Bench and MAGE --- the three component datasets of training and development sets for both English and multilingual subtasks.

\begin{table}[!ht]
\centering
\small
\begin{tabular}{ll|rrr|rrr}
\toprule
\multirow{2}{*}{Source} & \multirow{2}{*}{Sub-source} & \multicolumn{3}{c}{Training Set} & \multicolumn{3}{c}{Development Set} \\
\cmidrule(lr){3-5} \cmidrule(lr){6-8}
& & Human & Machine & Total & Human & Machine & Total \\
\midrule
& finance & 2579 & 3189 & 5768 & 1113 & 1301 & 2414 \\
& medicine & 886 & 883 & 1769 & 352 & 380 & 732 \\
HC3 & open\_g & 823 & 2339 & 3162 & 364 & 1015 & 1379 \\
& reddit\_tl5 & 34329 & 11680 & 46009 & 14781 & 4959 & 19740 \\
& wiki\_sai & 523 & 580 & 1103 & 245 & 262 & 507 \\
\midrule
& arxiv & 22484 & 30684 & 53168 & 9487 & 13003 & 22490 \\
& outfox & 2162 & 40973 & 43135 & 995 & 17390 & 18385 \\
M4GT-Bench & peerread & 3300 & 16169 & 19469 & 1398 & 6749 & 8147 \\
& reddit & 20353 & 32609 & 52962 & 8663 & 14076 & 22739 \\
& wikihow & 19454 & 35305 & 54759 & 8532 & 15168 & 23700 \\
& wikipedia & 19029 & 25341 & 44370 & 8145 & 10881 & 19026 \\
\midrule
& cmv & 6020 & 16592 & 22612 & 2618 & 7026 & 9644 \\
& cnn & 265 & 0 & 265 & 131 & 0 & 131 \\
& dialogsum & 210 & 0 & 210 & 98 & 0 & 98 \\
& eli5 & 15347 & 21849 & 37196 & 6451 & 9340 & 15791 \\
& hswag & 6806 & 19169 & 25975 & 2903 & 8085 & 10988 \\
& imdb & 269 & 0 & 269 & 107 & 0 & 107 \\
MAGE & pubmed & 273 & 0 & 273 & 105 & 0 & 105 \\
& roct & 6916 & 20008 & 26924 & 2930 & 8439 & 11369 \\
& sci\_en & 6613 & 14390 & 21003 & 2891 & 6145 & 9036 \\
& squad & 14519 & 14875 & 29394 & 6333 & 6330 & 12663 \\
& tldr & 5558 & 15808 & 21366 & 2329 & 6930 & 9259 \\
& wp & 7919 & 21215 & 29134 & 3393 & 9390 & 12783 \\
& xsum & 6992 & 22129 & 29121 & 2925 & 9621 & 12546 \\
& yelp & 25293 & 16058 & 41351 & 11039 & 6940 & 17979 \\
\midrule
Grand Total & & 228922 & 381845 & 610767 & 98328 & 163430 & 261758 \\
\bottomrule
\end{tabular}
\caption{\textbf{Monolingual subtask}: statistical information of training and development sets across different sources.}
\label{tab:english-traindev}
\end{table}

\begin{table}[!ht]
\centering
\small
\begin{tabular}{llccccccc}
\toprule
\multirow{2}{*}{Source} & \multirow{2}{*}{Sub-source} & \multirow{2}{*}{Lang} & \multicolumn{3}{c}{Training Set} & \multicolumn{3}{c}{Development Set} \\
\cmidrule(lr){4-6} \cmidrule(lr){7-9}
& & & Human & Machine & Total & Human & Machine & Total \\
\midrule
\multirow{11}{*}{HC3} & baike & zh & 2996 & 3211 & 6207 & 1247 & 1378 & 2625 \\
& \multirow{2}{*}{finance} & en & 2638 & 3135 & 5773 & 1054 & 1355 & 2409 \\
& & zh & 1103 & 1393 & 2496 & 438 & 560 & 998 \\
& law & zh & 494 & 353 & 847 & 196 & 145 & 341 \\
& \multirow{2}{*}{medicine} & en & 874 & 901 & 1775 & 364 & 362 & 726 \\
& & zh & 741 & 739 & 1480 & 317 & 327 & 644 \\
& nlpcc\_dbqa & zh & 1155 & 2718 & 3873 & 543 & 1094 & 1637 \\
& \multirow{2}{*}{open\_qa} & en & 840 & 2394 & 3234 & 347 & 960 & 1307 \\
& & zh & 5212 & 2683 & 7895 & 2148 & 1117 & 3265 \\
& psychology & zh & 3546 & 773 & 4319 & 1505 & 309 & 1814 \\
& reddit\_eli5 & en & 34510 & 11776 & 46286 & 14600 & 4863 & 19463 \\
& wiki\_csai & en & 546 & 594 & 1140 & 222 & 248 & 470 \\
\midrule
\multirow{13}{*}{M4GT-Bench} & Baike/Web QA & zh & 4068 & 4099 & 8167 & 1629 & 1819 & 3448 \\
& CHANGE-it NEWS & it & 0 & 4174 & 4174 & 0 & 1843 & 1843 \\
& \multirow{2}{*}{News/Wikipedia} & ar & 344 & 1770 & 2114 & 150 & 756 & 906 \\
& & de & 231 & 4462 & 4693 & 102 & 1957 & 2059 \\
& RuATD & ru & 684 & 630 & 1314 & 316 & 284 & 600 \\
& True \& Fake News & bg & 4205 & 3886 & 8091 & 1795 & 1694 & 3489 \\
& Urdu-news & ur & 2085 & 1676 & 3761 & 853 & 720 & 1573 \\
& arxiv & en & 22508 & 30649 & 53157 & 9463 & 13038 & 22501 \\
& id\_newspaper\_2018 & id & 1895 & 2081 & 3976 & 886 & 917 & 1803 \\
& outfox & en & 2196 & 40878 & 43074 & 961 & 17485 & 18446 \\
& peerread & en & 3291 & 16174 & 19465 & 1407 & 6744 & 8151 \\
& reddit & en & 20385 & 32535 & 52920 & 8631 & 14150 & 22781 \\
& wikihow & en & 19492 & 35187 & 54679 & 8494 & 15286 & 23780 \\
& wikipedia & en & 18975 & 25324 & 44299 & 8199 & 10898 & 19097 \\
\midrule
\multirow{14}{*}{MAGE} & cmv & en & 6009 & 16476 & 22485 & 2629 & 7142 & 9771 \\
& cnn & en & 275 & 0 & 275 & 121 & 0 & 121 \\
& dialogsum & en & 197 & 0 & 197 & 111 & 0 & 111 \\
& eli5 & en & 15214 & 21714 & 36928 & 6584 & 9475 & 16059 \\
& hswag & en & 6780 & 19163 & 25943 & 2929 & 8091 & 11020 \\
& imdb & en & 260 & 0 & 260 & 116 & 0 & 116 \\
& pubmed & en & 262 & 0 & 262 & 116 & 0 & 116 \\
& roct & en & 6820 & 19875 & 26695 & 3026 & 8572 & 11598 \\
& sci-gen & en & 6682 & 14308 & 20990 & 2822 & 6227 & 9049 \\
& squad & en & 14495 & 14914 & 29409 & 6357 & 6291 & 12648 \\
& tldr & en & 5526 & 15858 & 21384 & 2361 & 6880 & 9241 \\
& wp & en & 7941 & 21406 & 29347 & 3371 & 9199 & 12570 \\
& xsum & en & 6991 & 22202 & 29193 & 2926 & 9548 & 12474 \\
& yelp & en & 25502 & 16004 & 41506 & 10830 & 6994 & 17824 \\
\midrule
\multicolumn{3}{l}{Grand Total} & 257968 & 416115 & 674083 & 110166 & 178728 & 288894 \\
\bottomrule
\end{tabular}
\caption{\textbf{Multilingual subtask}: statistical information of training and development sets across different sources and languages.}
\label{tab:multi-traindev}
\end{table}

\clearpage
\begin{table}[!ht]
\centering
\small
\begin{tabular}{llrrrr}
\toprule
\multirow{2}{*}{Source} & \multirow{2}{*}{Model} & \multicolumn{2}{c}{Training Set} & \multicolumn{2}{c}{Development Set} \\
\cmidrule(lr){3-4}\cmidrule(lr){5-6}
& & Human & Machine & Human & Machine \\
\midrule
\multirow{9}{*}{\raggedright HC3} 
    & gpt-35 & 0 & 18671 & 0 & 7917 \\
    & human & 39140 & 0 & 16855 & 0 \\
    & bloomz & 0 & 21061 & 0 & 8991 \\
    & cohere & 0 & 20808 & 0 & 8896 \\
    & davinci & 0 & 19345 & 0 & 8210 \\
    & dolly & 0 & 8932 & 0 & 3931 \\
    & dolly-v2-12b & 0 & 1938 & 0 & 831 \\
    & gemma-7b-it & 0 & 12162 & 0 & 5240 \\
    & gemma2-9b-it & 0 & 8366 & 0 & 3629 \\
\midrule
\multirow{8}{*}{\raggedright M4GT-Bench} 
    & gpt-3.5-turbo & 0 & 25856 & 0 & 11005 \\
    & gpt4 & 0 & 9956 & 0 & 4300 \\
    & gpt4o & 0 & 10374 & 0 & 4247 \\
    & human & 86782 & 0 & 37220 & 0 \\
    & llama3-70b & 0 & 12333 & 0 & 5181 \\
    & llama3-8b & 0 & 12057 & 0 & 5290 \\
    & mixtral-8x7b & 0 & 15865 & 0 & 6623 \\
    & text-davinci-003 & 0 & 2028 & 0 & 893 \\
\midrule
\multirow{25}{*}{\raggedright MAGE} 
    & 13B & 0 & 5385 & 0 & 2367 \\
    & 30B & 0 & 5769 & 0 & 2380 \\
    & 65B & 0 & 5815 & 0 & 2404 \\
    & 7B & 0 & 5083 & 0 & 2166 \\
    & GLM130B & 0 & 4398 & 0 & 1842 \\
    & bloom$_{7b}$ & 0 & 5151 & 0 & 2201 \\
    & flan$_{5,\text{base}}$ & 0 & 6566 & 0 & 2887 \\
    & flan$_{5,\text{large}}$ & 0 & 6500 & 0 & 2893 \\
    & flan$_{5,\text{small}}$ & 0 & 6570 & 0 & 2811 \\
    & flan$_{5,\text{xl}}$ & 0 & 6429 & 0 & 2739 \\
    & flan$_{5,\text{xxl}}$ & 0 & 6532 & 0 & 2777 \\
    & gpt-3.5-turbo & 0 & 15991 & 0 & 6682 \\
    & gpt$_{j}$ & 0 & 3468 & 0 & 1480 \\
    & gpt$_{\text{neox}}$ & 0 & 4734 & 0 & 2021 \\
    & human & 103000 & 0 & 44253 & 0 \\
    & opt$_{1.3b}$ & 0 & 5553 & 0 & 2351 \\
    & opt$_{125m}$ & 0 & 5735 & 0 & 2469 \\
    & opt$_{3b}$ & 0 & 4988 & 0 & 2296 \\
    & opt$_{2.7b}$ & 0 & 5736 & 0 & 2586 \\
    & opt$_{30b}$ & 0 & 5637 & 0 & 2376 \\
    & opt$_{350m}$ & 0 & 5128 & 0 & 2252 \\
    & opt$_{6.7b}$ & 0 & 5642 & 0 & 2378 \\
    & opt$_{\text{iml}30b}$ & 0 & 6008 & 0 & 2619 \\
    & opt$_{\text{iml,max}1.3b}$ & 0 & 6176 & 0 & 2660 \\
    & t0$_{1b}$ & 0 & 6309 & 0 & 2620 \\
    & t0$_{3b}$ & 0 & 6602 & 0 & 2849 \\
    & text-davinci-002 & 0 & 14884 & 0 & 6359 \\
    & text-davinci-003 & 0 & 15304 & 0 & 6781 \\
\midrule
Grand Total & & 228922 & 381845 & 98328 & 163430 \\
\bottomrule
\end{tabular}
\caption{Generator distribution over three component of training and development sets.} 
\label{tab:generator-traindev-format}
\end{table}

\subsection{Test Sets}
\label{app:testset}
\tabref{tab:english_test_sets} shows the statistical distribution of English test sets in different domains and generators.
Tables \ref{tab:multilingual_test_sets_1} and \ref{tab:multilingual_test_sets_1} present the distribution of the multilingual test set over different languages, domains and generators (\href{https://docs.google.com/spreadsheets/d/1vIG10jO3QpA7tCMxBbLb1AVUvG0o5xosI9LDgmZSOG0/edit?usp=sharing}{see details}).

\begin{table}[ht]
\centering
\scalebox{0.65}{
\begin{tabular}{>{\raggedright\arraybackslash}p{3cm}|>{\centering\arraybackslash}p{2.5cm}>{\raggedleft\arraybackslash}p{1.5cm}>{\raggedleft\arraybackslash}p{1.5cm}>{\raggedright\arraybackslash}p{12.5cm}}
\toprule
\rule{0pt}{2.5ex}Source / Domain & License & \# Human & \# MGT & LLM Generator List \rule[-1.2ex]{0pt}{0pt}\\
\hline
\rule{0pt}{2.5ex}CUDRT-en subset & CC BY-SA 4.0 & 12939 & 10800 & GPT-3.5-turbo, Llama2, Llama3, ChatGLM, Baichuan, Qwen (1800 samples each)\\
\hline
\rule{0pt}{2.5ex}Mixset & CC BY-SA 4.0 & 600 & 3000 & -\rule[-1.2ex]{0pt}{0pt}\\
\hline
\rule{0pt}{2.5ex}LLM-DetectAlve-IELTS & huggingface & 1635 & 900 & llama-3.1-70B-versatile (900 samples)\rule[-1.2ex]{0pt}{0pt}\\
\hline
\rule{0pt}{2.5ex}IELTSDuck & Apache-2.0 & 10932 & 12418 & GPT-4o-mini-2024-07-18, (10932), llama-3.1-70B-versatile (1486)\rule[-1.2ex]{0pt}{0pt}\\
\hline
\rule{0pt}{2.5ex}NLPeer & Apache-2.0 & 5376 & 5376 & GPT-4o-2024-05-13 (5376)\rule[-1.2ex]{0pt}{0pt}\\
\hline
\rule{0pt}{2.5ex}Peersum & Github & 5157 & 6997 & GPT-4o-2024-08-06 (3501), GPT-4o-mini-2024-07-18 (3496)\rule[-1.2ex]{0pt}{0pt}\\
\hline
\rule{0pt}{2.5ex}Total & - & 36639 & 39491 & -\rule[-1.2ex]{0pt}{0pt}\\
\hline
\rule{0pt}{2.5ex}After deduplication & - & 35393 & 39363 & -\rule[-1.2ex]{0pt}{0pt}\\
\hline
\rule{0pt}{2.5ex}After removing short text & - & 34675 & 39266 & -\rule[-1.2ex]{0pt}{0pt}\\
\hline
\end{tabular}
}
\caption{Statistics of the English test set}
\label{tab:english_test_sets}
\end{table}
\begin{table}[ht]
\centering
\scalebox{0.65}{
\begin{tabular}{>{\raggedright\arraybackslash}p{3cm}|>{\centering\arraybackslash}p{2cm}>{\raggedleft\arraybackslash}p{1.5cm}>{\raggedleft\arraybackslash}p{1.5cm}>{\raggedright\arraybackslash}p{13cm}}
\hline
\rule{0pt}{2.5ex}Source / Domain & Language & \# Human & \# MGT & LLM Generator List \rule[-1.2ex]{0pt}{0pt}\\
\hline
\rule{0pt}{2.5ex}Cudrt-Subset & Chinese & 12565 & 1500 & GPT-3.5 (300), Qwen (300), GPT-4 (300), ChatGLM (300), Baichuan (300)\rule[-1.2ex]{0pt}{0pt}\\
\hline
\rule{0pt}{2.5ex}High School Student Essay & Chinese & 3502 & 1556 & GLM-4-9b-chat (778), Claude-3.5-sonnet (778)\rule[-1.2ex]{0pt}{0pt}\\
\hline
\rule{0pt}{2.5ex}Zhihu-Qa & Chinese & 12524 & 10269 & GPT-4o-2024-08-06 (3423), GPT-4o-mini-2024-07-18 (6846)\rule[-1.2ex]{0pt}{0pt}\\
\hline
\rule{0pt}{2.5ex}Mnbvc-Qa-Zhihu & Chinese & 3000 & 3000 & GPT-4o-2024-05-13 (3000)\rule[-1.2ex]{0pt}{0pt}\\
\hline
\rule{0pt}{2.5ex}Govreport & Chinese & 2975 & 17695 & GPT-4o-2024-05-13 (5932), ChatGLM3-6B (5821)\rule[-1.2ex]{0pt}{0pt}\\
\hline
\rule{0pt}{2.5ex}Easc (Summary) & Arabic & 153 & 306 & GPT-4o-2024-08-06 (306)\rule[-1.2ex]{0pt}{0pt}\\
\hline
\rule{0pt}{2.5ex}Tweets & Arabic & 1400 & 3400 & GPT-4 (1700), GPT-4o-2024-08-06 (1400), Qwen-2.5 72B (300)\rule[-1.2ex]{0pt}{0pt}\\
\hline
\rule{0pt}{2.5ex}Kalimat Youm 7 News & Arabic & 1000 & 2000 & GPT-4o-2024-05-13 (1000), Ace-GPT (1000)\rule[-1.2ex]{0pt}{0pt}\\
\hline
\rule{0pt}{2.5ex}Sanad (News) & Arabic & 3000 & 3000 & GPT-4o-2024-05-13 (3000)\rule[-1.2ex]{0pt}{0pt}\\
\hline
\rule{0pt}{2.5ex}Summaries & Russian & 6562 & 6582 & GPT-4o-2024-08-06 (3300), Vikhrmodels/Vikhr-Nemo-12B-Instruct-R-21-09-24 (3282)\rule[-1.2ex]{0pt}{0pt}\\
\hline
\rule{0pt}{2.5ex}News & Russian & 6494 & 6539 & GPT-4o-2024-08-06 (3295), Vikhrmodels/Vikhr-Nemo-12B-Instruct-R-21-09-24 (3244)\rule[-1.2ex]{0pt}{0pt}\\
\hline
\rule{0pt}{2.5ex}Wikipedia & Russian & 1025 & 3049 & GPT-4-0613 (999), Vikhrmodels/it-5.4-fp16-orpo-v2 (1025), AnatoliiPotapov/T-lite-instruct-0.1 (1025)\rule[-1.2ex]{0pt}{0pt}\\
\hline
\end{tabular}
}
\caption{Statistics of the multilingual test sets, part 1}
\label{tab:multilingual_test_sets_1}
\end{table}

\begin{table}[ht]
\centering
\scalebox{0.65}{
\begin{tabular}{>{\raggedright\arraybackslash}p{3cm}|>{\centering\arraybackslash}p{2cm}>{\raggedleft\arraybackslash}p{1.5cm}>{\raggedleft\arraybackslash}p{1.5cm}>{\raggedright\arraybackslash}p{13cm}}
\hline
\rule{0pt}{2.5ex}Source / Domain & Language & \# Human & \# MGT & LLM Generator List \rule[-1.2ex]{0pt}{0pt}\\
\hline
\rule{0pt}{2.5ex}Wikipedia & Hebrew & 1182 & 2173 & GPT-4-0613 (991), dicta-il/dictalm2.0-instruct (1182)\rule[-1.2ex]{0pt}{0pt}\\
\hline
\rule{0pt}{2.5ex}Wikipedia & German & 1865 & 2529 & GPT-4-0613 (957), LeoLM/leo-hessianai-13b-chat (1572)\rule[-1.2ex]{0pt}{0pt}\\
\hline
\rule{0pt}{2.5ex}Wikipedia & Norwegian & 1544 & 2543 & GPT-4-0613 (999), norallm/normistral-7b-warm-instruct (1544)\rule[-1.2ex]{0pt}{0pt}\\
\hline
\rule{0pt}{2.5ex}Wikipedia & Spanish & 600 & 600 & Llama 3.1 405B instruct (600)\rule[-1.2ex]{0pt}{0pt}\\
\hline
\rule{0pt}{2.5ex}Wikipedia & Dutch & 600 & 600 & Llama 3.1 405B instruct (600)\rule[-1.2ex]{0pt}{0pt}\\
\hline
\rule{0pt}{2.5ex}Wikipedia & kaz & 1300 & 1300 & GPT-4o-2024-08-06 (1300)\rule[-1.2ex]{0pt}{0pt}\\
\hline
\rule{0pt}{2.5ex}Dice (News) & Italian & 2800 & 2800 & Llama 3.1 405B instruct (2800)\rule[-1.2ex]{0pt}{0pt}\\
\hline
\rule{0pt}{2.5ex}News & Urdu & 13497 & 17472 & GPT-4o-2024-08-06 (17472)\rule[-1.2ex]{0pt}{0pt}\\
\hline
\rule{0pt}{2.5ex}News & Hindi & 600 & 600 & GPT-4o-2024-08-06 (600)\rule[-1.2ex]{0pt}{0pt}\\
\hline
\rule{0pt}{2.5ex}News & Japanese & 300 & 300 & GPT-4o-2024-08-06 (300)\rule[-1.2ex]{0pt}{0pt}\\
\hline
\rule{0pt}{2.5ex}News & Vietnamese & 600 & 600 & GPT-4o-2024-08-06 (600)\rule[-1.2ex]{0pt}{0pt}\\
\hline
\rule{0pt}{2.5ex}Wikipedia & Vietnamese & 600 & 600 & GPT-4o-2024-08-06 (600)\rule[-1.2ex]{0pt}{0pt}\\
\hline
\rule{0pt}{2.5ex}Poetry & Indonesian & 600 & 600 & GPT-4o-2024-08-06 (600)\rule[-1.2ex]{0pt}{0pt}\\
\hline
\rule{0pt}{2.5ex}Total & - & 80288 & 91613 & -\rule[-1.2ex]{0pt}{0pt}\\
\hline
\rule{0pt}{2.5ex}Non-duplicated & - & 78424 & 79305 & - \rule[-1.2ex]{0pt}{0pt}\\
\hline
\rule{0pt}{2.5ex}Remove Short Text & - & 73634 & 77791 & -\rule[-1.2ex]{0pt}{0pt}\\
\hline
\end{tabular}
}
\caption{Statistics of the multilingual test sets, part 2}
\label{tab:multilingual_test_sets_2}
\end{table}



\end{document}